\documentclass{article}

\usepackage{arxiv}

\usepackage[utf8]{inputenc} 
\usepackage[T1]{fontenc}    
\usepackage{hyperref}       
\usepackage{url}            
\usepackage{booktabs}       
\usepackage{amsfonts}       
\usepackage{nicefrac}       
\usepackage{microtype}      
\usepackage{lipsum}
\usepackage{graphicx}
\usepackage{amsmath}
\usepackage{amssymb}
\usepackage{animate}
\usepackage{algorithm} 
\usepackage{algpseudocode}
\usepackage{enumitem}
\graphicspath{ {./images/} }

\title{FloAt: \underline{Flo}w Warping of Self-\underline{At}tention for Clothing Animation Generation}

\author{
 Swasti S. Mishra \thanks{This work was done when the author was a part of Adobe Research.}\\
  University of Amsterdam\\
  \texttt{s.s.mishra@uva.nl} \\
   \And
 Kuldeep Kulkarni \\
  Adobe Research\\
  \texttt{kulkulka@adobe.com} \\
  \And
 Duygu Ceylan \\
  Adobe Research\\
  \texttt{ceylan@adobe.com} \\
  \And
 Balaji V. Srinivasan \\
  Adobe Research\\
  \texttt{balsrini@adobe.com} \\
}

\begin{document}
\maketitle

\begin{abstract}
   We propose a diffusion model-based approach, \emph{FloAtControlNet} to generate cinemagraphs composed of animations of human clothing. We focus on human clothing like dresses, skirts and pants. The input to our model is a text prompt depicting the type of clothing and the texture of clothing like leopard, striped, or plain, and a sequence of normal maps that capture the underlying animation that we desire in the output. The backbone of our method is a normal-map conditioned ControlNet ~\cite{zhang2023adding} which is operated in a training-free regime. The key observation is that the underlying animation is embedded in the flow of the normal maps. We utilize the flow thus obtained to manipulate the self-attention maps of appropriate layers. Specifically, the self-attention maps of a particular layer and frame are recomputed as a linear combination of itself and the self-attention maps of the same layer and the previous frame, warped by the flow on the normal maps of the two frames. We show that manipulating the self-attention maps greatly enhances the quality of the clothing animation, making it look more natural as well as suppressing the background artifacts. Through extensive experiments, we show that the method proposed beats all baselines both qualitatively in terms of visual results and user study. Specifically, our method is able to alleviate the background flickering that exists in other diffusion model-based baselines that we consider. In addition, we show that our method beats all baselines in terms of RMSE and PSNR computed using the input normal map sequences and the normal map sequences obtained from the output RGB frames. Further, we show that well-established evaluation metrics like LPIPS, SSIM, and CLIP scores that are generally for visual quality are not necessarily suitable for capturing the subtle motions in human clothing animations.
\end{abstract}

\begin{figure*}
    \centering	
    \animategraphics[autoplay,loop,width=0.95\textwidth, trim=3cm 0.4cm 3cm 1cm, clip]{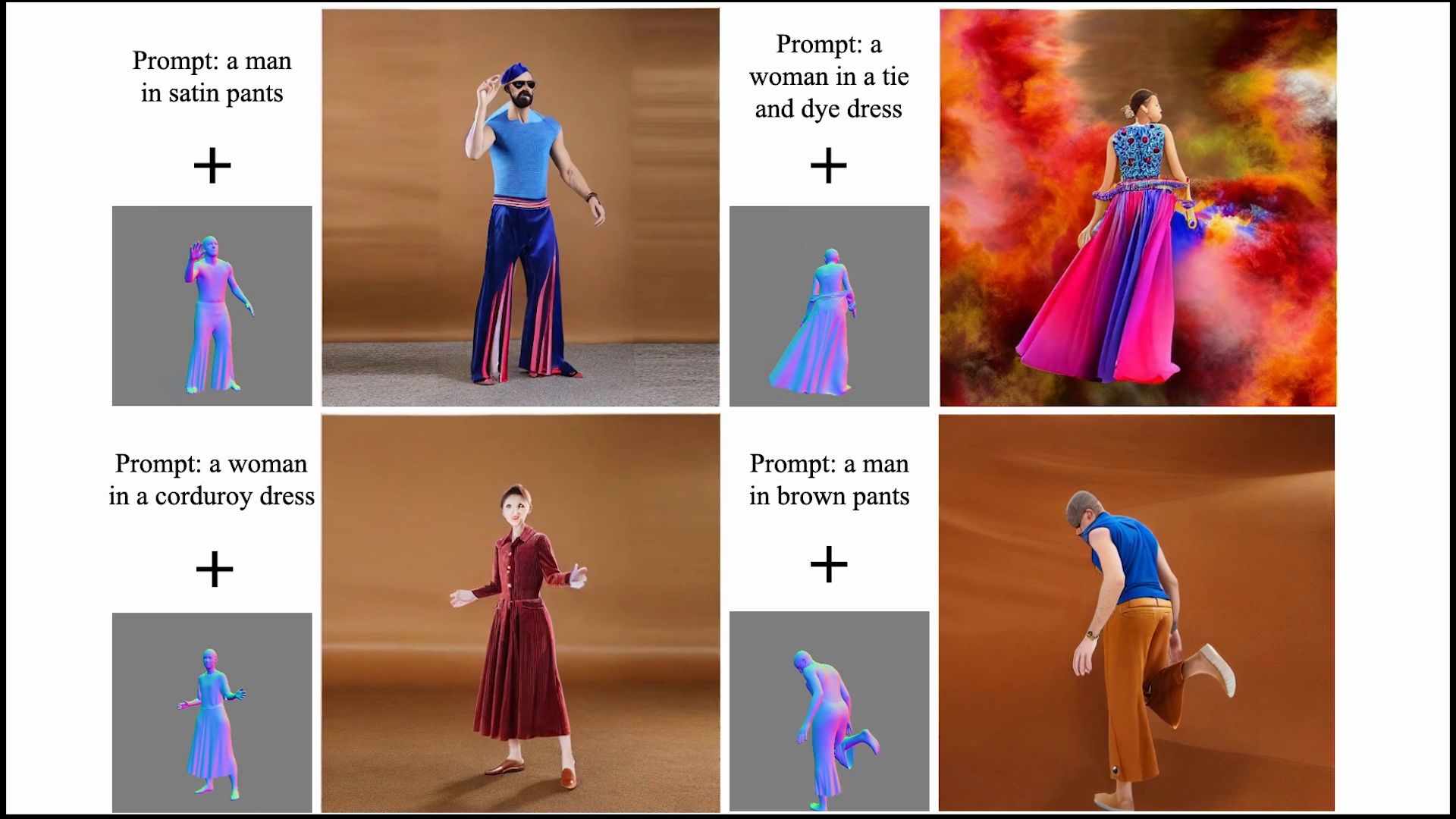}{fig/t_images/}{1}{20}
    \caption{We introduce a method for human clothing generation given a text prompt and a sequence of normal maps by manipulating the self-attention maps of normal-conditioned ControlNet using the flow information obtained from the normal maps. Our method is able to generate high-quality animation even for high-frequency textured dresses like stripes and tie and dye prints (see row 1). Please note that there are animations in the figure and are best viewed in Acrobat Reader.}
    \label{fig:teaser-fig}
\end{figure*}
\section{Introduction}
\label{sec:intro}
Animations of still images or cinemagraphs are engaging types of visual content that are becoming very popular with the advent of social media and digital advertising. People often prefer to post visuals that have some degree of motion to gain more traction on social media instead of a photo. Likewise, marketers are inclined to create animated photos of a product on online shopping sites to attract customers. However, very often this leads to the burden of having to capture the videos of the object that we want to be moving. Hence, there is a need to develop technologies to create these animations automatically from still photos \textit{ex-post facto}. The diversity in the type of objects that exist in nature, ranging from fluid elements like water, and fire to rigid bodies like the wheels of a car lends itself to a wide range of motion dynamics that makes it extremely difficult to describe the motion for all objects with just one model. Hence, researchers have focused on animating certain types of objects in their works like fluid elements \cite{Holynski_2021_CVPR, mahapatra2022controllable, mahapatra2023synthesizing}, oscillatory motion \cite{li2023generative}, repetitive elements \cite{halperin2021endless}, and garments \cite{bertiche2023blowing}. 
\\
We focus on generating cinemagraphs depicting animation of human clothing given input text prompts that describe the type of human clothing. Bertiche et al. \cite{bertiche2023blowing} show that a sequence of normal maps effectively describes the garment motion dynamics that involve geometry variations like folds and wrinkles owing to wind directions and is utilized to generate cinemagraphs of human clothing. Motivated by this, we develop a method that takes in a sequence of normal maps that captures the underlying animation of the human clothing in addition to the text prompt. Specifically, our method consists of a ControlNet backbone \cite{zhang2023adding} which is a normal map-conditioned diffusion model that is utilized to obtain a single RGB image. Applying ControlNet conditioned on normal maps for every frame separately, even with the same prompt leads to severe temporal inconsistency in garment texture as well as the background. We adopt the self-attention feature injection from pix2video \cite{ceylan2023pix2video} to obtain temporal consistency in the RGB frames thus obtained. While this helps us improve the garment motion, the animation is somewhat unnatural and the background flickering still remains. One way to circumvent these issues is by manipulating the self-attention maps of a particular layer and time step in a particular frame by utilizing the warped version of the self-attention maps from the previous frame. A concurrent work, LatentWarp \cite{bao2023latentwarp} that deals with a video-to-video problem obtains the optical flow from the input RGB frames and warps the latents from the previous frames at every time step. However, unlike LatentWarp, we do not have the luxury of having access to the RGB frames to be able to compute the flow. To circumvent this, we compute the flow on the \textit{normal maps} and show that the flow thus obtained has the information regarding the garment motion and can also be utilized to suppress the spurious background motion.
\\
More concretely, at every time step, we use the flow between the consecutive normal maps in the sequence to warp the self-attention map of the previous frame and manipulate the self-attention map of the current frame. We show that the flow on the normal maps is a very effective tool to make garment motion more natural as well as to alleviate the background motion to a great extent. Unlike the nearest GAN-based baseline CycleNet \cite{bertiche2023blowing}, we show that our method is able to generate the animations of high-frequency textured garments far more effectively and hence is more generalizable. The contributions of our paper are as follows:
\begin{itemize}[noitemsep,nolistsep]
    \item We propose a training-free and ControlNet-based method to generate cinemagraphs of human clothing given a text prompt and a sequence of normal maps.
    \item We show that flow on consecutive normal maps can be utilized to obtain temporal consistency and stability in the cinemagraphs through manipulation of the self-attention maps. 
    \item We show that our method generates a wide variety of pleasing animations of high-frequency textures in clothing, unlike any other previous method (figure \ref{fig:teaser-fig}).
    \item We show qualitatively and quantitatively that our method is far superior to all baselines, diffusion model based or otherwise.
\end{itemize}

\section{Related Work}
\label{sec:rel_work}
\textbf{Single Image Animation:}
Single-image animation is a reasonably well-studied problem in the context of fluid elements like water, smoke and fire. There exist a number of approaches that perform quite well. Specifically, Holysinki et al. \cite{Holynski_2021_CVPR} propose two generative adversarial network (GAN) \cite{goodfellow2014generative} models to animate a still image. The first model computes a single optical flow describing the motion and the second model takes in the input image and the flow to compute the subsequent frames. Mahapatra et al. \cite{mahapatra2022controllable} improve on this to provide the user control through a small number of arrow directions that are converted into a dense flow by exponential functions. This allows the user to generate several animations from the same still image. However, both of these methods \cite{Holynski_2021_CVPR, mahapatra2022controllable} suffer from the drawback that they allow the background in the fluid region to move. To address this issue, Fan et al. \cite{fan2023simulating} devise a method to decompose the pixel space into two layers, an unchangeable background and a fluid region and only animate the fluid region using similar flow-based GAN methods. There have been other works \cite{chuang2005animating, halperin2021endless, li2023generative} that allow for the animation of generic elements. Chuang et al. \cite{chuang2005animating} divide the animatable elements into different layers and apply stochastic motion texture to each of these elements separately and then composite them back into a single animation. Halperin et al. \cite{halperin2021endless} exploit the repeating patterns in elements like staircase, and giant wheel to obtain self-similarity descriptions that are utilized to obtain animations. These methods are not suitable for human clothing animation that we desire in this work, as animation involves changes in the geometry of the clothing. Closest to our work is CycleNet \cite{bertiche2023blowing} which animates human clothing and other garments from a single still image. The approach involves expressing the input image into normal map space and then obtaining the normal map sequence that describes the animation of the garment using a GAN. CycleNet uses an intrinsic image decomposition approach to generate the final RGB images which is an ill-posed problem, especially in the case of garments with high-frequency texture patterns. We take inspiration from this method and also utilize normal map sequences to drive the human clothing animation. However, unlike CycleNet, we leverage pre-trained image diffusion models, in particular, normal map conditioned ControlNet \cite{zhang2023adding} to synthesize the RGB images. We inject the flow computed from normal maps into the self-attention layers of the diffusion model. Owing to the robust ControlNet backbone and flow injection, our method is able to generate high-quality animations of high-frequency textures in clothing, whereas CycleNet can produce good-quality animations of plain garments but fails in generating textured clothing animations. 
\\
\textbf{Diffusion Model based Video Stylization:} After the enormous success of diffusion models for text-to-image generation \cite{ho2020denoising, song2020denoising, rombach2022high}, there have been several attempts to imbibe them for video generation and manipulation problems, in particular the video stylization problem because of the training-free regime that diffusion models offer to operate in, unlike the GANs. Specifically, researchers have adopted the base text-to-image stable diffusion model \cite{rombach2022high} to tackle video translation. Approaches like pix2video \cite{ceylan2023pix2video}, Tune-A-Video \cite{zhangjie2022tune}, Rerender-a-Video\cite{yang2023rerender}, Video-p2p \cite{liu2023video}, Fatezero\cite{qi2023fatezero} and Text2video\cite{khachatryan2023text2video} all operate in a training-free regime and inflate self-attention using features from anchor frames or previous frames. However, the temporal consistency achieved may not always be up to the desired extent. A recent line of concurrent works \cite{geyer2023tokenflow, cong2023flatten, bao2023latentwarp, wu2023lamp}  focus on injecting flow during the generation process to enforce better temporal consistency. TokenFlow \cite{geyer2023tokenflow} leverages the cross-attention maps to obtain correspondences for the same token across frames and the correspondences are used to warp the attention maps. LatentWarp \cite{bao2023latentwarp} computes the optical flow explicitly on the input RGB frames and uses them to manipulate the latent at every time step. Our method utilizes the benefits of the self-attention feature injection similar to pix2video\cite{ceylan2023pix2video} that are subsequently warped using the flow computed on the input normal maps. Mahapatra et al. \cite{mahapatra2023synthesizing} propose a method, Eulergraphs to obtain stylistic cinemagraphs of fluid elements from text prompts. However, unlike our method which is only based on diffusion models, this method leverages the diffusion model to generate the first frame and then leverages GAN models to generate the frames of the animation. The use of flow/correspondences to enhance temporal coherency has been explored in recent works, Rerender-A-Video ~\cite{yang2023rerender} and TokenFlow~\cite{geyer2023tokenflow}. We distinguish our work from Rerender-a-video ~\cite{yang2023rerender} in that the latter introduces the usage of optical flow computed on the input RGB frames (that we do not have access to in our problem) for manipulation of the latent itself while our method introduces the use of flow computed on surface normal maps, instead, for the manipulation of self-attention features. We distinguish our work from TokenFlow~\cite{geyer2023tokenflow} in that the method requires RGB frames as input to compute the DDIM versions. In this process, the self-attention maps are retained and used to compute nearest-neighbour fields. Adapting TokenFlow for our problem by DDIM inverting surface normal maps resulted in undesirable frames that were devoid of colour and texture, and in general, looked like noisy versions of the surface normal maps hence we do not compare with this method.
\section{Methodology}
\label{sec:method}
\begin{figure*}[hbt!]
    \centering
    \includegraphics[width=\textwidth]{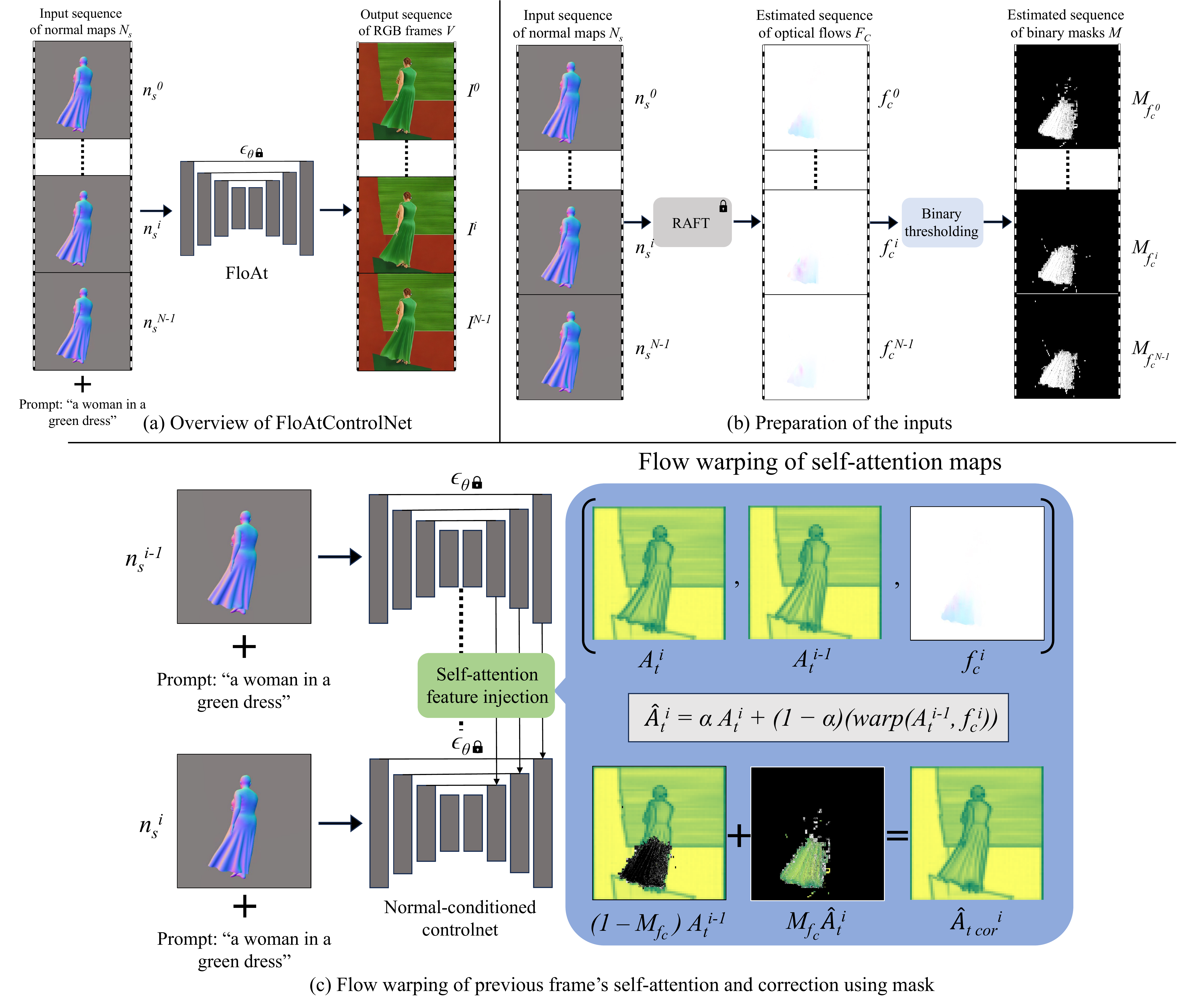}
    \caption{\textbf{Overview of FloAtControlNet.} Given a text prompt and an input sequence of normal maps, we first compute the flow over the sequence of normal maps using the RAFT~\cite{teed2020raft} model and threshold it as mentioned to get a binary mask. We sequentially input the normal maps and text prompt into the normal-conditioned ControlNet~\cite{zhang2023adding}. Next, we perform self-attention feature injection inspired by Pix2Video~\cite{ceylan2023pix2video} to ensure temporal consistency of the visual features of the generated sequence. Further, during the denoising process, we recompute the self-attention map for a particular frame as a linear combination of itself and the flow-warped corresponding self-attention map from the previous frame to suppress the spurious motions in the generation. Finally, to eliminate the background flickering artifacts we do self-attention feature correction as stated in Equation~\ref{eq:attn_cor}.}
    \label{fig:pipeline}
\end{figure*}

We aim to do text-to-cinemagraph generation conditioned on the sequence of normal maps, $N_{s} = \{n_{s}^i\}_{i = 0}^{N-1}$ and a text prompt $\tau$. As illustrated in Figure~\ref{fig:pipeline}, we propose a training-free method for realistic human clothing animation generation. We leverage the normal-conditioned ControlNet~\cite{zhang2023adding} model by injecting cross-frame self-attention features to make the generated video temporally coherent. We then effectively manipulate the self-attention maps by using the optical flow $F_{C}$ obtained from $N_{s}$ to suppress any undesirable motion in the generated sequence.

\subsection{Preliminaries: Stable Diffusion and ControlNet}



\textbf{Stable Diffusion (SD):} Stable Diffusion models~\cite{rombach2022high} are currently the most widely used text-to-image generation framework that operates in a latent image space. A pretrained encoder on RGB images is utilized to map images into the latent space, then this latent representation is progressively denoised to obtain an intermediate representation, and finally, high-resolution images are generated from the intermediate representation using a decoder. The network architecture for denoising is constructed using a U-Net $\epsilon_{\theta}(x_{t}, t, \tau)$ which incorporates residual, self-attention, and cross-attention blocks. For a given denoising time step $t$, text prompt $\tau$ and a latent representation $x_{t}$, $\epsilon_{\theta}(x_{t}, t, \tau)$ predicts the noise to be subtracted from $x_{t}$, in order to get a high-quality output image after the decoding step. During the inference step of the diffusion model, the SD starts from a random Gaussian noise $x_{T}\sim\mathcal{N}(0,I)$ and progressively denoises it to get $x_{0}$. To delve deeper, the residual block involves convolving activations from the preceding layer, while the cross-attention block controls features based on the provided text prompt. Within the self-attention block, features are projected into $d$-dimensional queries $Q$, keys $K$, and values $V$. The output of a self-attention block at denoising time step $t$ is given by:

\begin{equation}
    A_{t} = Softmax(\frac{Q_{t}K_{t}^{T}}{\sqrt{d}})V_{t}
\end{equation}


\textbf{ControlNet:} ControlNet~\cite{zhang2023adding} is a stable diffusion model based variant that allows the user to provide additional controls like normal maps, depth maps, edge maps and pose. Denoting the $c_{f}$ as the extra condition, the noise prediction generated by U-Net with ControlNet is represented as $\epsilon_{\theta}(x_{t}, t, \tau, c_{f})$. Given a text prompt $\tau$ and a surface normal map $n_{s}$, we utilize the normal-conditioned ControlNet in a training-free manner to obtain individual frames from the normal maps.

\subsection{Baseline: Self-Attention Feature Injection in ControlNet}
\label{sec:baseline}
Given a sequence of normal maps $N_{s}$ and a text prompt $\tau$, we generate a video sequence $V = \{I^{i}\}_{i=0}^{N-1}$. Each $I^{i}$ is generated by denoising a latent vector $x_{T}^{i}$ for $T$ time steps to get $x_{0}^{i}$ which is then decoded through the decoder. In order to generate a frame sequence that is closely related through a subtle motion but has the same object characteristics, we generate each $I^{i}$ from the same noise initialization $x_{T}$, i.e.,  $x_{T} = x_{T}^{i}  \forall i \in [0, N-1]$. However, as we show (in the supplemental) this fails to achieve the desired motion characteristics even though the high-level textures of the clothing and background are retained. Therefore, we incorporate cross-frame self-attention feature injection inspired by Pix2Video~\cite{ceylan2023pix2video} to ensure better temporal consistency. This refers to utilizing input features, $h$ from an anchor frame, $0^{th}$ and the previous frame to compute $K$ and $V$ for the current frame (refer to line 1 in the Algo~\ref{alg:algo1}). We call this baseline \textbf{`FeatInControlNet'} where we concatenate the self-attention features of an anchor frame and a previous frame when generating the current frame.

If the output of a self-attention block at denoising time step $t$ is given by:
  $A_{t} = Softmax(\frac{Q_{t}K_{t}^{T}}{\sqrt{d}})V_{t}$,
where $Q_t \in 1 \times 64 \times 64 \times 320$, $K_{t} \in 1 \times 64 \times 64 \times 320$, $V_t \in 1 \times 64 \times 64 \times 320$ and hence, $A_t \in 1 \times 64 \times 64 \times 320$, cross-frame self-attention feature injection refers to utilizing input features, $h$ from an anchor frame, $0^{th}$ and the previous frame to compute $K$ and $V$ for the current frame (refer to line 1 in the Algo~\ref{alg:algo1}). We call this baseline \textbf{`FeatInControlNet'} where we concatenate the self-attention features of an anchor frame and a previous frame when generating the current frame.

\subsection{Visualising Self-Attention and Flow on Normal Maps}
\label{sec:attn_viz}
\begin{figure*}[hbt!]
    \centering
    \includegraphics[width=\textwidth]{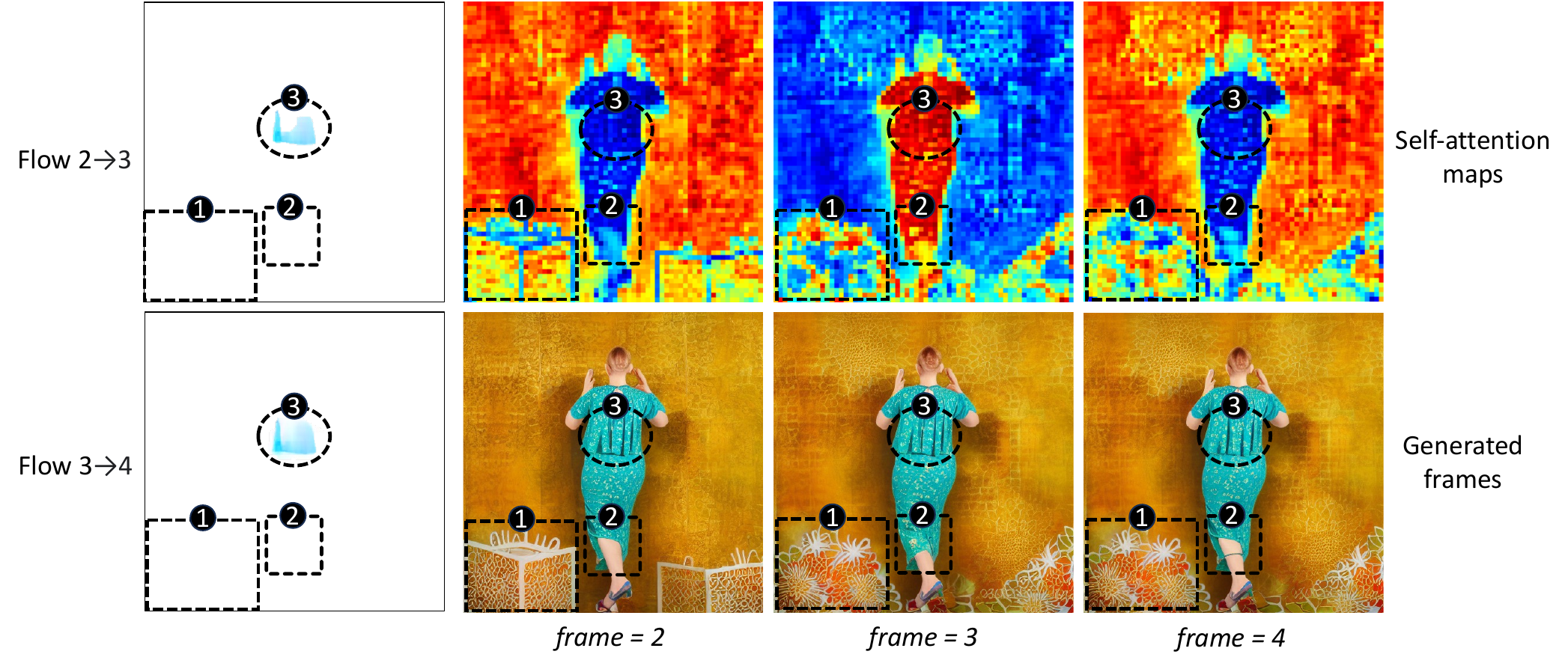}
    \caption{\textbf{Self-attention visualisation.} For a given \textbf{FeatInControlNet} generated sequence, we take the first PCA component of the self-attention map for frames 2, 3 and 4, at the last layer of the 3rd ConvUpBlock of the U-Net and plot its heatmap at the final denoising step. We mark 3 spatial regions of the self-attention maps, depicted as dotted bounding boxes (1, 2) and an ellipse (3) and its corresponding region in the generated frame. In the first column, we show the flow maps computed on frames 2 to 3 (top left) and frames 3 to 4 (bottom left). We also specify the mentioned spatial regions on the flow maps. We observe that even in the regions where there is zero flow (bounding boxes 1 \& 2), the self-attention maps change noticeably. This results in undesirable motion in the generated sequence. This forms the motivation for the flow injection into the self-attention maps.}
    \label{fig:attn_vis}
\end{figure*}

We inspect the results of \textbf{FeatInControlNet} to find that it has flickering artifacts and undesirable and somewhat unnatural animation in the generations (see Figure~\ref{fig:attn_vis}, row 2). To understand this phenomenon, we visualize the self-attention maps for each frame $i$ at denoising time step $t$ by taking the most dominant PCA \cite{mackiewicz1993principal} component of $A_{t}^{i}$ and visualizing it as a heatmap. We compute this for all the self-attention layers of all the convolution up-blocks of the U-Net. We show that the last layer of the 3rd ConvUpBlock has the maximum influence on the final generated image and is thus highly correlated with the motion present in the normal map (in supplement). Through Figure~\ref{fig:attn_vis} we demonstrate that the self-attention maps of \textbf{FeatInControlNet} have flickering artifacts that percolate into the final generated frame. We visualize the self-attention map from the last layer of the 3rd ConvUpBlock at the final denoising step. We can clearly see the structural similarity in the corresponding spatial regions of the visualized self-attention maps (row 1 in Figure~\ref{fig:attn_vis})  and the generated images (row 2 in Figure~\ref{fig:attn_vis}). It is easy to notice that the change in the generated frame is a consequence of the undesired change in the self-attention maps, thus resulting in artifacts as exhibited by bounding boxes, 1 and 2. We conclude from this study that it is vital to connect the self-attention maps across frames through external information in order to restrict the motion only to the desired regions. We choose to compute the flow, $F_{C}$ on the normal maps, $N_{s}$ and show that it can be utilized to manipulate the self-attention maps. Due to the lack of any input video or input flow, we utilize a pretrained RAFT~\cite{teed2020raft} to compute the flow on normal maps. To this end, we treat the input normal maps (3-channels) as if they are RGB images and feed normal maps for consecutive frames into the RAFT model to obtain the flow.

\subsection{FloAtControlNet: Flow Warping of Self-Attention Maps}
\label{sec:float}
Given a sequence of normal maps, $N_{s}$ we have a sequence of flows computed from the normal maps, $F_{C} = \{f_{c}^{i}\}_{i = 0}^{N-1}$. As shown above, the changes in the self-attention maps across frames are not commensurate with the flow information obtained from the normal maps. In order to incorporate the flow information into the self-attention maps, we manipulate the self-attention map for frame $i$ at denoising time step $t$, $\hat{A_{t}^{i}}$ (we drop the layer and block notation for simplicity, and as mentioned in Section~\ref{sec:attn_viz}). Specifically, for a given frame, we recompute the self-attention maps as a linear combination of itself and the flow-warped version of the self-attention map of the previous frame. The manipulated is mathematically in equation \ref{eq:flow_warp}
\begin{equation}
\label{eq:flow_warp}
    \hat{A_{t}^{i}} = \alpha A_{t}^{i} + (1 - \alpha) (warp(A_{t}^{i-1},f_{c}^{i})),
\end{equation}
where, $\alpha$ is a scalar constant governing the linear combination, $warp(.)$ is the bilinear interpolation function used to apply the flow between the $(i-1)^{th}$ and $i^{th}$ frames and $\hat{A_{t}^{i}}$ is the recomputed self-attention map for denoising step $t$ and frame $i$. The exact value of $\alpha$ is chosen to be $0.4$ determined through a simple ablation as shown in the supplementary. Further, we improve the self-attention maps by explicitly enforcing the non-motion regions of the self-attention maps to remain constant. We achieve this by first computing a binary mask by thresholding the flow, $f_{c}^{i}$. Using this mask, we compute the final attention map, $\hat{A_{t}^{i}}_{cor}$ by borrowing the self-attention features from the previous frame at the same denoising step, for the spatial regions corresponding to motion (or zero flow) less than the threshold and retaining the features in the regions above the threshold to the manipulated attention map, $\hat{A_{t}^{i}}$ as given in equation \ref{eq:attn_cor}. 
\begin{equation}
\label{eq:attn_cor}
    \hat{A_{t}^{i}}_{cor} = (1 - M_{f_{c}})A_{t}^{i-1} + M_{f_{c}}\hat{A_{t}^{i}}
\end{equation}
where $M_{f_{c}}$ is a binary mask obtained by thresholding the flow, $f_{c}^{i}$.
\begin{equation}
\label{eq:mask_comp}
    M_{f_{c}}(x, y) =
    \begin{cases}
      1, & \text{if}\ f_{c}(x, y) \geq threshold \\
      0, & \text{otherwise}
    \end{cases}
\end{equation}
We term our method of infusing the flow information into the self-attention maps as \textbf{`FloAtControlNet'}. To summarize, our method to manipulate self-attention maps in the form of an algorithm for a given time step, $t$ and frames, $i=1,..,N-1$ for the last layer of the 3rd ConvUpBlock in the U-Net is given below. The for loop (lines 3-6) in the Algorithm~\ref{alg:algo1} is an expanded version of equations \ref{eq:flow_warp}. and \ref{eq:attn_cor}. Please note that the implementation incorporates matrix operations and the algorithm below is simplified only for explanation purposes.
\begin{algorithm}
	\caption{FloAt} 
	\begin{algorithmic}[1]
          \For {$i=1,..,$N-1}
          \State{
            $Q_{t}^{i} = W^{q}h_t^{i}$; $K_{t}^{i} = W^{k} [h_t^{0}, h_t^{i-1}]$ ;
            $V_{t}^{i} = W^{v} [h_t^{0}, h_t^{i-1}]$}
            \State{
            $A_t^i = \text{Softmax}(\frac{Q_{t}^{i}(K_{t}^{i})^{T}}{\sqrt{d}})V_{t}^{i}$ }
            \For {$s=0,1,..,319$}
           \State{ $\hat{A_{t}^{i}}(:,:,:,s) = \alpha A_{t}^{i}(:,:,:,s) + (1 - \alpha) (\text{warp}(A_{t}^{i-1}(:,:,:,s),f_{c}^{i}))$}
           \State{ $\hat{A_{t}^{i}}_{cor}(:,:,:,s) = (1 - M_{f_{c}})A_{t}^{i-1}(:,:,:,s) + M_{f_{c}}\hat{A_{t}^{i}}(:,:,:,s)$ }
           \EndFor
         \State{
            $A_t^i = \hat{A_{t}^{i}}_{cor}$}
         \EndFor
	\end{algorithmic} 
\label{alg:algo1}
\end{algorithm}

\textbf{Background Suppression:} The flow in the background obtained from the normals is zero. By setting the flow $f_{c}^{i}$ to zero in line 4 of the algorithm, we obtain $\hat{A_{t}^{i}} = \alpha A_{t}^{i} + (1 - \alpha) (A_{t}^{i-1})$. However, given that $M_{f_{c}}$ is obtained by thresholding $f_{c}^{i}$, it is zero in the background regions. Therefore, for the background region, $A_{t_{cor}}^{i} = A_{t}^{i-1}$, for all $i$. Thus by induction, we have  $\hat{A_{t}^{i}}_{cor} = A_{t}^{0}$ for all regions in the background. This ensures that background noise is nearly filtered out in the generated RGB sequence. Through our experiments in Section~\ref{sec:exp} we show \textbf{FloAtControlNet} significantly improves the generation quality and outperforms the other baselines in both quantitative and qualitative aspects.

\section{Experiments}
\label{sec:exp}
\textbf{Dataset and Set-up:} In order to evaluate the efficacy of our method to generate cinemagraphs of human clothing from text prompts, we create a dataset of $2990$ pairs of normal map sequences ($230$) and text prompts ($13$). The normal map sequences are randomly selected from the Cloth 3D dataset \cite{bertiche2020cloth3d} simulated under the effect of wind~\cite{bertiche2023blowing}. 
The selection of text prompts is carefully curated to encompass a diverse range of both plain and textured clothing designs across different types of wear like dresses, pants and skirts, resulting in a rich array of generated output frames. Typical examples of text prompts are `Women in a plain red dress', and `A man in blue spotted pants'. The exhaustive list of employed text prompts and the normal map sequence pairs are available in the supplemental. Our study includes extensive experimentation with both baseline models and our proposed approach, revealing the superior performance of our method across specified metrics and in terms of visual quality. For our experiments across different baselines, we use the same latent noise vector for all frames in a given example. We use the default parameters for the classifier-free guidance and the number of inference steps as specified in the ControlNet~\cite{zhang2023adding} paper. However, for ControlVideo~\cite{zhang2023controlvideo} we use the parameters chosen by the authors.
\subsection{Baselines}
\textbf{CycleNet Reshading:} We establish a baseline derived from CycleNet~\cite{bertiche2023blowing} by using an input image generated by the normal-conditioned ControlNet model for a specified text prompt. This input image is produced using the same initial latent noise vector used in our other experiments. Subsequently, we employ the CycleNet model to generate a cinemagraph by combining the aforementioned image with a sequence of normal maps provided as input.
\\
\textbf{ControlNet:} We establish an additional baseline by utilizing a normal conditioned ControlNet~\cite{zhang2023adding}. This approach maintains a consistent text prompt and latent noise vector across all frames, with the only variation being the sequence of normal maps.
\\
\textbf{Rerender-A-Video Adapt:} A close baseline for our work is Rerender-A-Video ~\cite{yang2023rerender} that was originally proposed for video-to-video stylization task. We adapt the shape-aware latent warp component from this work to create a baseline, `Rerender-A-Video Adapt'.
\\
\textbf{ControlVideo:} We compare our work to ConrtolVideo~\cite{zhang2023controlvideo} which has a ControlNet backbone but is proposed for video-to-video translation task. Here we adapt it to our problem by changing the structure guidance model to the normal conditioned ControlNet~\cite{zhang2023adding} and directly pass the sequence of normal maps along with the text prompt as input. 
\\
\textbf{FeatInControlNet:} We compare our method with FeatInControlNet, where we use the normal conditioned ControlNet with self-attention feature injection as specified in Section~\ref{sec:baseline}. Here, we concatenate the self-attention features of an anchor frame (which in our case is the starting frame) and the previous frame during the generation of the current frame. We perform this for the first 50\% of the denoising steps as specified in Pix2Video~\cite{ceylan2023pix2video}.
\\
\textbf{FeatInControlNet+Mask:} We devise a baseline that uses not only the mask component of our method but also the flow warping. More concretely, we first start by doing the self-attention map correction using the mask $M_{f_{c}}$ during each denoising step for the corresponding frames for FeatInControlNet. We call this baseline `FeatInControlNet+Mask'. The corrected self-attention map is obtained similar to Equation~\ref{eq:attn_cor} as follows:
\begin{equation}
    A_{t_{cor}}^{i} = (1 - M_{f_{c}}) A_{t}^{i-1} + M_{f_{c}} A_{t}^{i}
\end{equation}
\subsection{Evaluation Metrics}
We evaluate our method and baselines on a set of pairs of synthetic surface normal sequences and text prompts for both normal map conditioning as well as the visual quality of the animation. 

\textbf{Metrics for Normal Conditioning:}
First, we estimate the normal maps from the generated video sequence using an off-the-shelf normal estimator~\cite{bae2021estimating} and obtain the root of mean squared error (RMSE), and peak signal-to-noise ratio (PSNR) between the input sequence of normal maps and the estimated sequence of normal maps and call them N-RMSE and N-PSNR respectively. Further, we compute the flow on both the input and estimated normal maps using ~\cite{teed2020raft} and use them to compute the RMSE (F-RMSE) and PSNR (F-PSNR) between them. 

\begin{table}
    \centering
    \caption{Quantitative evaluation of the normal map conditioning for various methods. Our method is better than the closest baseline, Rerender-A-Video Adapt showing that flow warping in self-attention map space yields better results than in latent space. Our method is significantly better than CycleNet Reshading which does not always generate animations that respect the normal map sequence.}
    \begin{tabular}{|l|c|c|c|c|}
        \hline
         Methods & N-RMSE $\downarrow$ & N-PSNR $\uparrow$ & F-RMSE $\downarrow$ & F-PSNR $\uparrow$\\
         \hline
         CycleNet Reshading ~\cite{bertiche2023blowing} & 27.853 & 11.069 & 66.357 & 9.774\\
         ControlNet & 19.553 & 13.647 & 65.843 & 9.712 \\
         FeatInControlNet & 19.488 & 13.708 & 59.769 & 10.637\\
         Rerender-A-Video Adapt ~\cite{yang2023rerender} & 19.456 & 13.674 & 56.873 & 10.988\\
         FloAtControlNet (Ours) & \textbf{19.396} & \textbf{13.748} & \textbf{54.887} & \textbf{11.455}\\
         \hline
    \end{tabular}
    \label{tab:norm_quant}
\end{table}

\textbf{Metrics for Visual Quality:}
We do not have the ground truth for the generated video sequence and hence, it is impossible to compute the SSIM~\cite{wang2004image} and LPIPS~\cite{zhang2018unreasonable} scores directly. To circumvent this issue, we select $k$ equally spaced frames from the generated RGB sequence, compute the LPIPS and SSIM scores of all other frames with respect to these $k$ frames individually and then compute the average of the LPIPS and SSIM scores thus obtained. We call these metrics, Self-LPIPS and Self-SSIM. We hypothesise that since the motion in the sequence of normal maps is only restricted to the clothing region and is quite subtle, any changes in the structural and perceptual characteristics of the clothing and background are captured through these average LPIPS and SSIM metrics. 

We also report the CLIP semantic (CLIP\textsubscript{s}) and CLIP consistency (CLIP\textsubscript{c}) scores to assess the similarity between text embedding and the frame generated and the inter-frame semantic similarities. However, these metrics are more suitable for video-to-video translation tasks, where both the foreground and the background regions can change notably from one frame to another. In our case owing to the subtle motion and practically no change in the object characteristics from one frame to another, we show that CLIP scores are not necessarily the right measures to evaluate animations such as the ones considered in our paper. 

\begin{figure*}[hbt!]
    \centering
    \includegraphics[width=\textwidth]{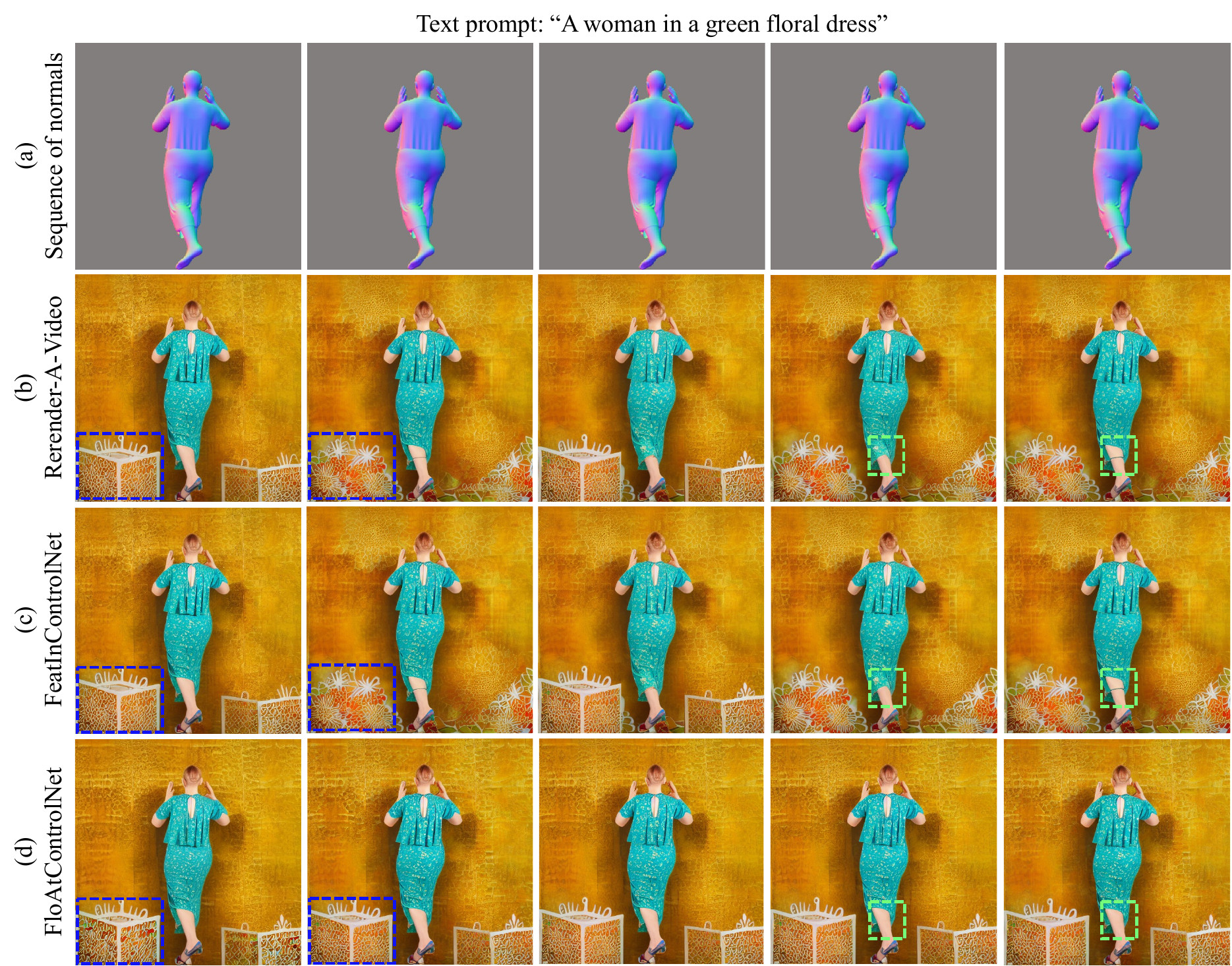}
    \caption{Qualitative results for our different methods and Rerender-A-Video Adapt~\cite{yang2023rerender}. Our approach is able to generate a temporally coherent sequence of frames, by suppressing artifacts in the no-motion region as demarcated by the dotted bounding boxes. Rerender-A-Video fails to suppress the spurious motion in the background.}
    \label{fig:qual2}
\end{figure*}

\begin{table}
\centering
\caption{Quantitative evaluation of the visual coherence of the different methods. The proposed methods for visual quality, Self-LPIPS, Self-SSIM, CLIP\textsubscript{s} and CLIP\textsubscript{c} fail to capture the subtle motions as they focus mainly on the high-level object and scene characteristics.}
    \begin{tabular}{|l|c|c|c|c|}
        \hline
        Methods & Self-LPIPS $\downarrow$ & Self-SSIM $\uparrow$ & CLIP\textsubscript{s} $\uparrow$ & CLIP\textsubscript{c} $\uparrow$ \\
        \hline
        CycleNet Reshading ~\cite{bertiche2023blowing} & \textbf{0.002} & \textbf{0.997} & 28.044 & \textbf{99.997} \\
        ControlNet & 0.242 & 0.704 & 28.064 & 97.693 \\
        FeatInControlNet & 0.073 & 0.901 & \textbf{28.113} & 99.682 \\
        ControlVideo & 0.048 & 0.925 & 27.954 & 99.739 \\
        Rerender-A-Video Adapt ~\cite{yang2023rerender} & 0.070 & 0.915 & 26.791 & 99.833 \\
        FloAtControlNet (Ours) & 0.025 & 0.956 & 28.108 &	99.893 \\
        \hline
    \end{tabular}
\label{tab:rgb_quant}
\end{table}

\begin{figure*}[hbt!]
    \centering
    \includegraphics[width=\textwidth]{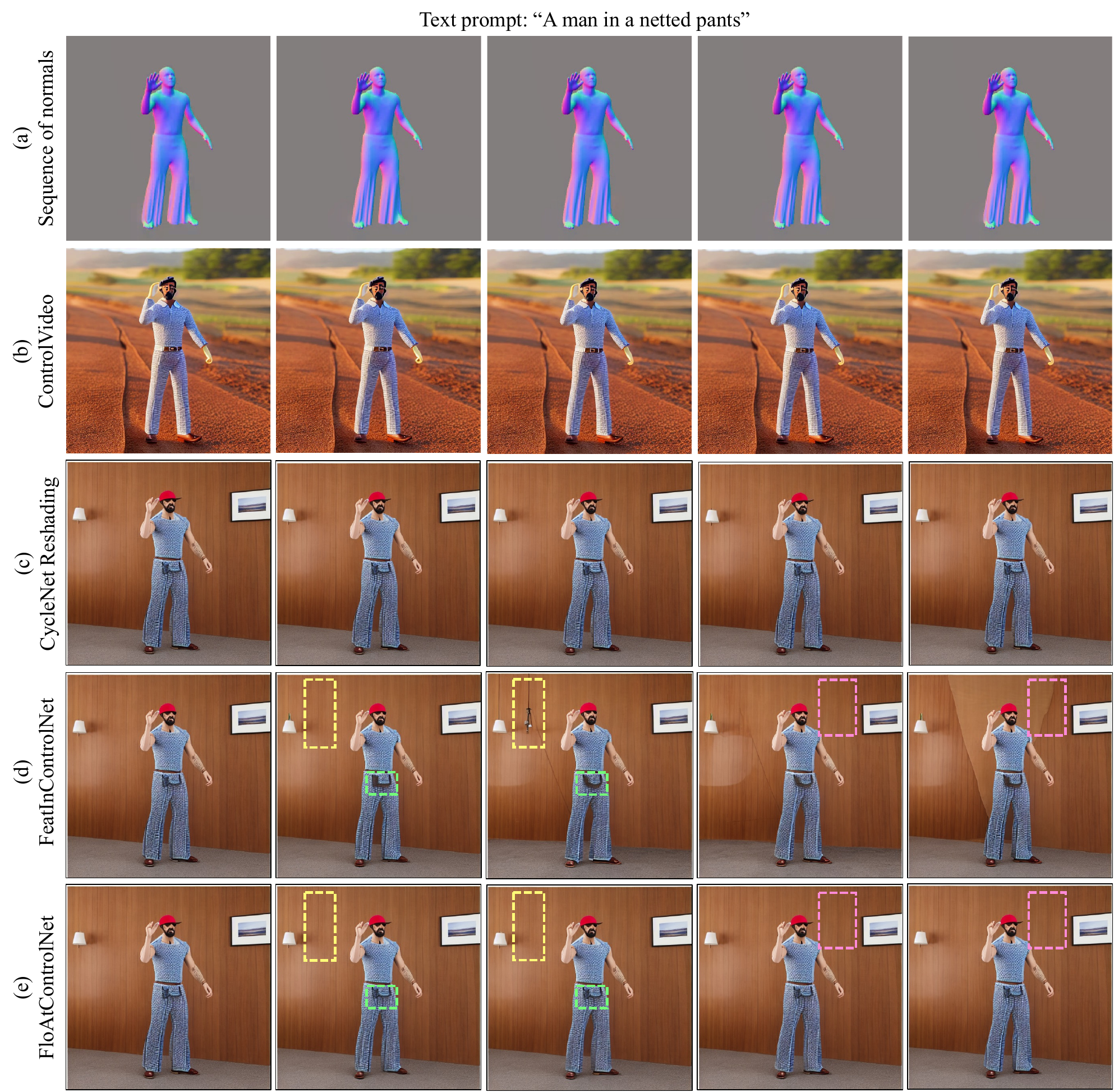}
    \caption{Qualitative results for our different methods and CycleNet Reshading ~\cite{bertiche2023blowing}, ControlVideo~\cite{zhang2023controlvideo}. Our approach is able to generate a temporally coherent sequence of frames, by suppressing artifacts in the no-motion region as demarcated by the dotted bounding boxes. CycleNet and ControlVideo fail to generate intricate motion present in the input normal sequences.}
    \label{fig:qual1}
\end{figure*}

\subsection{Results}
\textbf{Quantitative Comparisons:}
Table~\ref{tab:norm_quant} shows that our method outperforms all baselines in terms of all metrics, N-RMSE, N-PSNR, F-RMSE, F-PSNR proposed to evaluate the effectiveness of normal conditioning. In particular, our method performs significantly better than CycleNet Reshading ~\cite{bertiche2023blowing}, the previous baseline for normal conditioned cinemagraph generation. We also highlight that our method performs better than Rerender-A-Video Adapt thus showing the flow warping of the self-attention maps is a better approach than warping the latents. In Table~\ref{tab:rgb_quant} we report the Self-LPIPS, Self-SSIM, CLIP\textsubscript{s} and CLIP\textsubscript{c} scores with averaging $k=10$ frames. The minimal differences across the baselines demand a more nuanced look. Unlike video-to-video translation tasks, our animations involve subtle motion with virtually no change in object characteristics from one frame to another. All the metrics considered for visual quality assessment in this table are designed to capture substantial changes in both foreground and background regions, which do not occur in subtle motions that we consider. 


\textbf{Qualitative Comparisons:} The qualitative results for our different methods and the baselines can be found in figures~\ref{fig:qual2} \& ~\ref{fig:qual1} (see supplementary for animations). We show that our approach is able to generate a temporally coherent sequence of frames, by suppressing artifacts in the no-motion region as marked by the dotted bounding boxes. FloAtControlNet provides the most natural clothing animation (see supplementary) as well as alleviating the artifacts in the background as indicated by blue bounding boxes in Figure~\ref{fig:qual1} and in green and yellow bounding boxes in Figure~\ref{fig:qual2}. For Renderer-A-Video, the contents of the blue bounding box in the second column in Figure~\ref{fig:qual2} indicate a signficantly changing background, whereas for our method, FloatControlNet, the change in the background is greatly alleviated. Similarly for Render-A-video, the last column in Figure ~\ref{fig:qual2} clearly show artifact in the green dress, whereas for our method, the artifact is absent. This shows that flow warping in the self-feature map space is better than flow warping in the latent space as is the case in Renderer-A-Video.
In figure~\ref{fig:qual1}, it is clear that our method, FloatControlNet is able to generate intricate that is present in the pants region of the normal map sequence. However, CycleNet reshading and ControlVideo fails to generate any motion. Similarly, the background motion for our method is greatly suppressed as compared to FeatControlNet due to the flow warping of the self-attention maps. 

\begin{figure}
    \centering
    \includegraphics[width=\columnwidth]{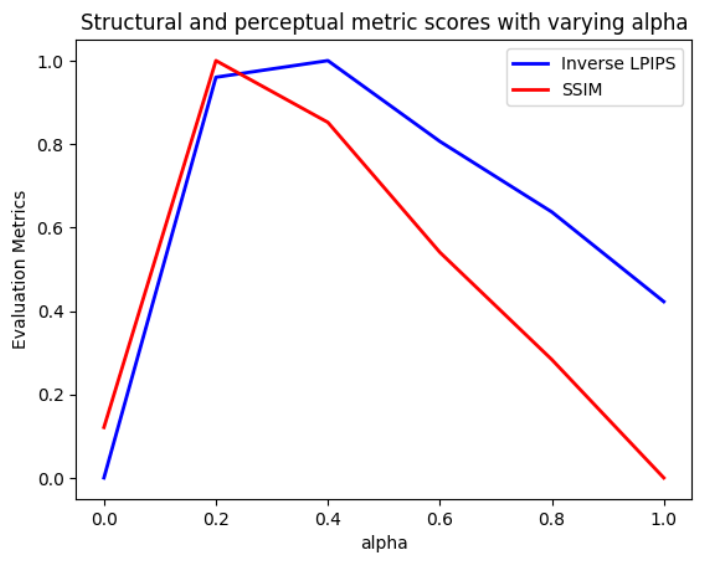}
    \caption{We report the min-max normalized inverse Self-LPIPS and Self-SSIM scores with the specified $\alpha$ values. This helps us determine the optimal weightage to be given to the self-attention of the current frame over the flow-warped self-attention of the previous frame. For all our experiments, we fix $\alpha$ for our experiments to be 0.4.}
    \label{fig:alpha_var}
\end{figure}

\textbf{Ablations:}
To analyze the effectiveness of the design choices, we test additional baselines which are variants of our method. As shown in Table~\ref{tab:ablate}, FeatInControlNet+Mask outperforms FeatInControlNet on all the structural and perceptual metrics and equally well or better on the scores computed on the normal maps. However, we notice that the animation of the clothing still looks a bit unrealistic. Therefore, we add the flow warping of the self-attention maps (Section~\ref{sec:float}) to enhance the perceptual results further to get FloAtControlNet. As shown in Table~\ref{tab:ablate}, our method significantly outperforms FeatInControlNet+Mask in the structural and perceptual metrics and performs equally well on the metrics computed on the normal maps. We also vary the $\alpha$ parameter in Equation~\ref{eq:flow_warp} to find the value that gives the best results. The results generations for the perceptual metrics, when we vary the $\alpha$, is shown in Figure~\ref{fig:alpha_var}. 
\begin{table}
    \centering
    \begin{tabular}{|l|c|c|}
    \hline
       Methods  & Self-LPIPS $\downarrow$ & Self-SSIM $\uparrow$ \\
       \hline
       FeatInControlNet & 0.073 & 0.901\\
       FeatInControlNet+Mask & 0.041 & 0.930 \\
       FloAtControlNet (Ours) & \textbf{0.026} & \textbf{0.956} \\
       PCAFloAtControlNet & 0.030 & 0.946\\
       \hline
    \end{tabular}
    \caption{Quantitative evaluation of the different methods tested for ablations. As shown in this table, FloAtControlNet beats all baselines both in terms of Self-LPIPS and Self-SSIM metrics, thus showcasing much higher-quality animations.}
    \label{tab:ablate}
\end{table}

\textbf{User Study:}
Owing to the limitations of the quantitative metrics, we augment our evaluation with a qualitative user study. We present randomly sampled animations generated by CycleNet Reshade, Rerender-A-Video Adapt, and FloAtControlNet from \textit{10 examples} to \textit{67 users} to select their preference for the best animation. In this study, we show the input sequence of normal maps (Input Video) and ask the following question to the user: \textit{In which of the following generated videos does the animation in the ``clothing region'' best resemble the ``clothing region'' animation shown in the Input Video?} and provide the CycleNet Reshade, Rerender-A-Video Adapt, and FloAtControlNet generated animations as the options. The findings demonstrate that FloAtControlNet (our method) was the preference of \textbf{41.3\%} of the users, compared to CycleNet Reshade and Rerender-A-Video Adapt which were the preference of 27.3\% and 31.3\% of the users respectively, establishing our method as the most favoured.

\section{Limitations and Discussions:}

We presented a training-free ControlNet-based approach for human clothing cinemagraphs from text prompts and a sequence of normal maps depicting the animation. Specifically, we proposed the use of flow on normal maps to manipulate the self-attention maps and show that such a manipulation of features improves the animation quality greatly and suppresses the background motion. Unlike previous methods, our method has the ability to generate cinemagraphs of high-frequency textured clothing and not just plain clothing. Through qualitative and quantitative experiments, we showed that our method FloAtControlNet beats all baselines by a significant margin, thus corroborating the need for infusing the flow on normal maps. One of the limitations of our work is that it requires high-quality normal map sequences both for conditioning and flow computation. Erroneous estimates of normal maps used as inputs can lead to severe imperfections in the animations. While one can estimate normal map sequences from a single image like CycleNet, erroneous estimates can lead to severe imperfections in the animations. 
\\
We observe that all diffusion model-based methods that we consider are an improvement over the previous method, CycleNet. The absence of motion in certain cases is primarily attributed to  CycleNet Reshading stage that relies on intrinsic image decomposition. 
The shading does change in response to variations in the input normal directions, ensuring a dynamic representation of the scene's lighting conditions. However, the challenge arises during the final step of generating the RGB video. CycleNet Reshade composites the shading map sequence with the original reflectance map to produce the ultimate RGB video output. Importantly, this process only alters the shading without actually warping the texture of the garment or scene. This sometimes results in static regions in the high-frequency texture regions. This observation is consistent with the limitations pointed out in the CycleNet work \cite{bertiche2023blowing}.
\\
Another limitation of our paper is that the self-attention maps that are manipulated by the flow are of the size $64 \times 64$ which is a far cry from the size of the frames generated by our method that is $512 \times 512$. While our method greatly suppresses motion in the background motion where there is no flow in the normal maps as compared to all baselines, sometime undesirable artifacts may remain in the background regions where there is no flow owing to the 8X upsampling by the decoder of the diffusion model. Such undesirable artifacts are reminiscent of those seen in the wider problem of image editing using diffusion models wherein the regions that are not intended to be edited also change to some extent \cite{kawar2023imagic}. It is a problem generally not observed in GANs as the editing is done in the native resolution. However, the flow injection to the self-attention improves the temporal consistency and motion quality in the foreground regions.

\section{Supplementary Materials}

All the visual results can be accessed through the \href{https://swastishreya.github.io/float/}{https://swastishreya.github.io/float}. Specifically, there are five links in the page given below:
\\
\textbf{Link 1}: Flow Warping on Different Layers of the Self-Attention Maps \\
\textbf{Link 2}: Attention Maps Comparison Videos  \\
\textbf{Link 3}: Qualitative Comparison Videos \\
\textbf{Link 4}: Applications of Our Method \\
\textbf{Link 5}: Sample of Normal Map Sequences Chosen for Evaluation \\
\textbf{Link 6}: TokenFlow Generations

\subsection{Dataset -- Link 5}
\label{sec:dataset}
We evaluate our methods and baselines qualitatively and quantitatively on a dataset of 2990 (normal, prompt) unique pairs of normal map sequences from the wind-simulated Cloth 3D dataset~\cite{bertiche2020cloth3d} (230) and the following list of text prompts (13):
\\
1. `a woman/man in red dress/pants'\\
2. `a woman/man in brown dress/pants'\\
3. `a woman/man in green dress/pants'\\
4. `a woman/man in white dress/pants'\\
5. `a woman/man in polka dotted dress/pants'\\
6. `a woman/man in floral dress/pants'\\
7. `a woman/man in leopard print dress/pants'\\
8. `a woman/man in netted dress/pants'\\
9. `a woman/man in corduroy dress/pants'\\
10. `a woman/man in lace dress/pants'\\
11. `a woman/man in satin dress/pants'\\
12. `a woman/man in striped dress/pants'\\
13. `a woman/man in tie and dye dress/pants' \\

Our dataset comprises 230 normal map sequences (a sample given in \textbf{Link 5}), encompassing examples featuring both men and women. Consequently, we tailor the text prompts for our dataset pairs based on the corresponding normal map sequence. For instance, when the normal map sequence seems to depict a man, we employ the `a man in * pants' variation of the text prompt. Conversely, in the case of a woman, we utilize the `a woman in * dress' variation, where * denotes the texture of the clothing.

\subsection{Experimentation with Attention Layers -- Links 1 and 2}
\label{sec:attn_layers}

To identify the self-attention layer with the most significant impact on output generation, we conduct flow warping independently on layers 1, 2, and 3 of the 3rd ConvUpBlock within the UNet. We note that applying flow warping to the self-attention maps of the earlier ConvUpBlocks has no discernible effect on the final generation. This outcome is expected, given that the fine-grained structural details only manifest in the self-attention layers of the last ConvUpBlock. Consequently, our analysis focuses on manipulating the self-attention layers of the 3rd ConvUpBlock.

The \textbf{Link 1} illustrates an example with generations while manipulating layers 1, 2, and 3 independently. When applying flow warping to the first layer, flickering issues persist (e.g., noticeable changes in sleeves across frames). Similarly, the application of flow warping to the second layer results in pronounced artifacts affecting the clothing in the animation region. In contrast, when flow warping is applied to the last self-attention layer, a notable improvement in generation quality is observed, and flickering issues are alleviated.

In \textbf{Link 2} we demonstrate how our method based on flow warping of the self-attention maps is able to enhance the quality of the animations. We visualize the self-attention maps of the FeatInControlNet method and its corresponding generations, as well as the self-attention maps of FloAtControlNet and its respective generations. As mentioned in Section 3.3 and elucidated in Figure 3 of the main paper we visualize the sequence of PCA of the self-attention maps as frames advance. It is clear that the flickering artifacts in the self-attention maps do get corrected after the application of flow warping of the self-attention maps. From the RGB frames shown in the link before and after flow-warping, it is evident that the correction of the self-attention maps is reflected in the higher-quality animations generated.

\subsection{Qualitative Results -- Link 3}
\label{sec:qual2}

We include the qualitative result videos in \textbf{Link 3}. These results provide a comparative analysis of several methods, including \textbf{ControlNet}, \textbf{ControlVideo}, \textbf{CycleNet Reshade}, \textbf{Rerender-A-Video}, \textbf{FeatInControlNet}, and \textbf{FloAtControlNet}. Additionally, we supply the sequence of normal maps used alongside the input text prompt for reference.

The \textbf{ControlNet} method employs the same starting latent noise vector for all frames. However, as demonstrated in the results, frames generated in this manner lack temporal coherence. 

In the case of the \textbf{ControlVideo} method, the outputs reveal minimal to no motion in the generated videos. In instances where motion is observed, the generations are significantly impacted by flickering artifacts.

The \textbf{CycleNet Reshade} method, on the other hand, produces satisfactory outputs for examples with subtle motion (e.g., examples 1, 3, and 8). However, in other cases, it either fails to generate any motion (e.g., examples 2, 9, and 11) or generates unrealistic clothing animation (e.g., examples 4, 6, and 10). 

The \textbf{Rerender-A-Video} method, which incorporates the shape-aware latent warp component from the work of Yang et al. \cite{yang2023rerender}, yields decent outcomes by mitigating background motion in certain scenarios (e.g., examples 2 and 6). Nevertheless, it struggles to curb the background and undesirable motion in other instances (e.g., examples 4 -- background and 5 -- sleeves). Moreover, in the majority of cases, the animation in the clothing region appears excessively abrupt (e.g., examples 3, 6, and 9).

We additionally juxtapose our own baseline, \textbf{FeatInControlNet}, against our proposed method. While it demonstrates commendable performance in generating clothing animation across the majority of cases (e.g., examples 2, 3, and 9), it falls short in mitigating extraneous motion in the background (e.g., examples 2, 6, and 10) and addressing flickering in the clothing region, particularly where no motion is present in the corresponding spatial region of the normal map sequence (e.g., observe the sleeves in examples 5 and 12).

Overall, our method, \textbf{FloAtControlNet} demonstrates superior qualitative performance by effectively addressing flickering artifacts and producing more realistic animations.

\subsection{Applications -- Link 4}
\label{sec:app}

Our method can be employed in conjunction with the CycleNet model~\cite{bertiche2023blowing} to generate clothing animations for high-frequency textured dresses, addressing cases where CycleNet encounters difficulties. Given an input image generated using a text prompt, we can predict the sequence of normal maps using the \textbf{Wind Cyclic UNet} from CycleNet. Subsequently, our method is applied to generate the final cinemagraph using the predicted sequence of normals.

The results are illustrated in \textbf{Link 4} where we present a comparison between our method and CycleNet. The examples include scenarios with high-frequency textured dresses and highly reflective clothing materials such as satin (first table -- Applications of Our Method). Additionally, we showcase comparisons for generations featuring changing wind directions, where the sequence of normal maps is predicted by CycleNet (second table -- Animations with Changing Wind Directions). These results illustrate that our method can effectively complement existing approaches to generate a wide variety of animations for the same input image.

\subsection{TokenFlow Results -- Link 6}

We employ TokenFlow for our task and present some examples in \textbf{Link 6}. We input the sequence of normal maps as the source video and give the text prompt corresponding to the texture that is desired in the clothing. As mentioned in Section 2, we distinguish our work from TokenFlow~\cite{geyer2023tokenflow} in that the method requires RGB frames as input to compute the DDIM versions. In this process, the self-attention maps are retained and used to compute nearest-neighbour fields. 
As depicted in \textbf{Link 6}, DDIM inverting surface normal maps occasionally results in undesirable frames that are devoid of colour and texture (e.g., examples 2 and 3). Crucially, across all instances, the temporal inconsistency of the clothing texture throughout the video makes the animation look ``jumpy''. Consequently, we refrain from comparing this method with our own.

\bibliographystyle{unsrt}

\end{document}